\documentclass[10pt,twocolumn,letterpaper]{article}

\usepackage[pagenumbers]{cvpr}      
\definecolor{cvprblue}{rgb}{0.21,0.49,0.74}
\newcommand{\best}[1]{\cellcolor{cvprblue!30}\textbf{#1}}
\newcommand{\secondbest}[1]{\cellcolor{cvprblue!15}\underline{#1}}
\usepackage[pagebackref,breaklinks,colorlinks,allcolors=cvprblue]{hyperref}
\usepackage{algorithm}
\usepackage{algorithmic}
\usepackage{amsmath}
\usepackage{colortbl}
\usepackage{newfloat}
\usepackage{listings}
\usepackage{tabularx}
\usepackage{multirow}
\usepackage{svg}
\usepackage{wrapfig}
\newsavebox{\fullcelltablebox}
\NewDocumentEnvironment{fullcelltabular}{m m m +b}{%
  \setlength{\tabcolsep}{0pt}%
  \sbox{\fullcelltablebox}{\begin{tabular}{#3}#4\end{tabular}}%
  \setlength{\tabcolsep}{\dimexpr(#1-\wd\fullcelltablebox)/#2/2\relax}%
  \begin{tabular*}{#1}{#3}#4\end{tabular*}%
}{}

\def\confName{CVPR}
\def\confYear{2027}

\title{FreqForcing: Autoregressive Long Video Generation \\ via Spectral Self-Anchoring}

\author{
  \textbf{Jiatong Li$^{1}$,\enspace Yulun Zhang$^{1}$\footnotemark[1],\enspace Leo Liang$^{2}$,\enspace Linghe Kong$^{1}$\footnotemark[1]}\thanks{Corresponding authors: Yulun Zhang, yulun100@gmail.com; Linghe Kong, linghe.kong@sjtu.edu.cn} \\
  \textsuperscript{1}Shanghai Jiao Tong University,\\
  \textsuperscript{2}Tencent HY Team
}

\makeatletter
\apptocmd{\@maketitle}{%
  \par\noindent
  \begin{minipage}{0.9\textwidth}
    \centering
    \includegraphics[width=\linewidth]{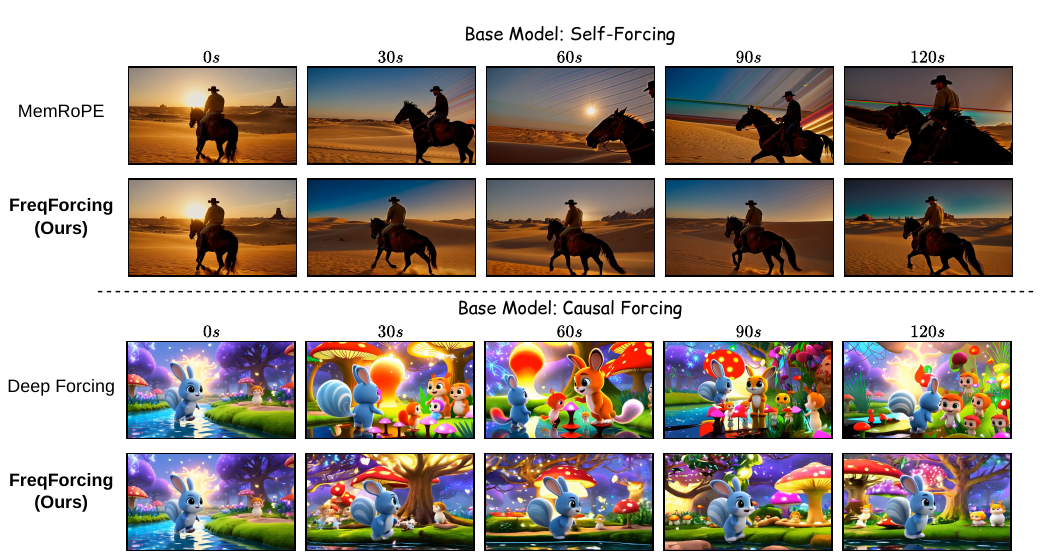}
    \vspace{-6.5mm}
    \captionof{figure}{
      We propose FreqForcing, a \textbf{training-free} framework that enables AR diffusion models trained on short video clips (\eg Self-Forcing~\cite{huang2026self} and Causal Forcing~\cite{zhu2026causal}) to generate minutes-long videos. FreqForcing significantly improves color consistency and motion richness over long horizons.
    }
    \label{fig:teaser}
  \end{minipage}
  \par\vspace{1em}
}{}{%
  \PackageError{teaser}{Failed to patch title block}{}
}
\makeatother
\begin{document}
\maketitle

\begin{abstract}
Autoregressive video diffusion models enable real-time streaming video generation. However, errors introduced during self-rollout accumulate over long horizons, manifesting as color drift, motion stagnation, and eventual visual collapse. Existing training-free methods attribute error accumulation to the loss of historical context and use attention sink to compensate for this loss. However, we find that visual degradation primarily stems from erroneous historical tokens, whose key-value pairs distort attention outputs. Although attention sink partially alleviates this issue, visual degradation persists. Motivated by the above analysis, we propose \textbf{FreqForcing}, a training-free framework that rectifies attention outputs in long-video generation via Spectral Self-Anchoring (SSA). The proposed SSA leverages the low-frequency components of anchor attention to maintain long-horizon visual stability, while preserving recent changes through the high-frequency components of local attention. FreqForcing extends models pretrained on \textit{5s} clips to minutes-long generation without finetuning. Extensive experiments show that FreqForcing outperforms existing training-free methods quantitatively and qualitatively.
\end{abstract}

\section{Introduction}
Video diffusion models have advanced rapidly, with large-scale diffusion transformers and video foundation models such as Sora~\cite{videoworldsimulators2024}, HunyuanVideo~\cite{wu2025hunyuanvideo}, and Wan~\cite{wan2025wan} synthesizing realistic, temporally coherent, and instruction-following videos. However, applications like interactive world models~\cite{bruce2024genie} and embodied decision-making~\cite{yang2023unisim} demand streaming frame generation with low latency, high throughput, and online controllability.

Autoregressive (AR) video diffusion models~\cite{yin2025slow,huang2026self} offer an efficient alternative by generating frames sequentially while reusing cached key-value pairs. In particular, Self-Forcing~\cite{huang2026self} performs AR self-rollout during training to bridge the train-test gap, enabling real-time streaming generation with competitive quality. However, the causal attention that enables streaming generation also conditions each chunk on imperfect histories, so errors are repeatedly fed back and accumulate over long horizons.

Existing remedies fall into two groups. Training-based methods~\cite{liu2025rolling,zhu2026causal,cui2025self} improve long-horizon stability through better distillation strategies, but they are computationally demanding. Training-free methods instead intervene at inference time, either improving RoPE~\cite{yesiltepe2026infinity,yi2025deep,li2026train} or adopting more effective KV cache management strategies~\cite{yi2025deep,li2026rollingsink,kim2026memrope} to preserve historical context. However, these methods still cannot guarantee the visual quality when generating minute-long videos.

\begin{figure*}[h]
\centering
\includegraphics[width=\linewidth]{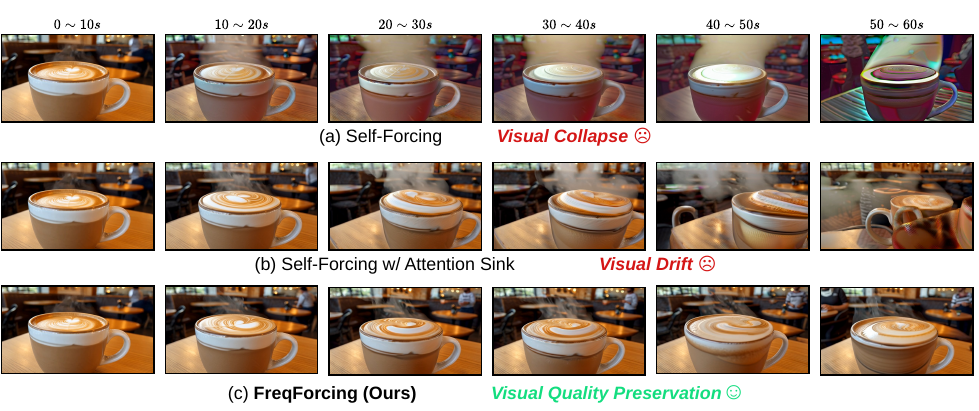}
\vspace{-6.8mm}
\caption{\textbf{Visual comparison.} \textbf{(a)} Self-Forcing~\cite{huang2026self} suffers from severe color drift and visual fading. \textbf{(b)} Self-Forcing with attention sink partially mitigates these artifacts but still exhibits visual drift. \textbf{(c)} FreqForcing produces visually stable videos with spectral self-anchoring.}
\label{fig:spectral_drift}
\end{figure*}

To better understand the error accumulation, we conduct visual comparisons and a spectral analysis. As shown in Figs.~\ref{fig:spectral_drift}(a) and \ref{fig:error_analysis}(b), error accumulation manifests as color drift and flickering in pixel space. These artifacts are consistent with a decline in the spatial low-frequency components and a rise in the temporal high-frequency components.

Recent work has shown that attention sink is effective in both LLM inference~\cite{ghadia2025dialogue,shilacache,xiao2024efficient} and AR video generation~\cite{liu2025rolling,yang2025longlive}. Therefore, we also measure the effect of attention sink. As shown in Figs.~\ref{fig:spectral_drift}(b) and \ref{fig:error_analysis}(b), adding attention sink substantially improves color consistency. In the spectral analysis, both the decline of spatial low-frequency components and the rise of temporal high-frequency components become less pronounced.

To understand how attention sink improves long video generation, we track the energy of attention outputs. As shown in Fig.~\ref{fig:error_analysis}(c), the energy of multiple attention heads grows progressively during generation with vanilla Self-Forcing~\cite{huang2026self}. Inference with attention sink slows this growth, yet attention output energy continues to rise and visual quality still degrades. These observations suggest that rectifying attention output could help preserve visual quality over long generation horizons.

Motivated by prior works~\cite{lu2024freelong,lu2025freelong++,chen2026freespec} and the above analysis, we propose \textbf{FreqForcing}, a training-free framework for long-video generation via Spectral Self-Anchoring (SSA). Specifically, SSA employs two complementary attention branches during inference: a local branch preserves recent changes and an anchor branch provides stable guidance. Their outputs are subsequently fused in the frequency domain to mitigate degradation and enhance long-term consistency. As illustrated in Fig.~\ref{fig:teaser}, FreqForcing achieves superior subject and background consistency compared to baseline methods. Extensive quantitative and qualitative results demonstrate that FreqForcing extends Self-Forcing~\cite{huang2026self} and Causal Forcing~\cite{zhu2026causal}, originally pretrained on 5-second videos, to minutes-long generation, while maintaining high visual quality and temporal consistency.

Our contributions are summarized as follows:
\begin{itemize}
    \item We conduct a comprehensive analysis of video quality degradation during AR generation. We illustrate how errors accumulate and affect the quality of videos.
    \item We propose \textbf{FreqForcing}, a training-free framework that mitigates error accumulation through \textbf{Spectral Self-Anchoring (SSA)}. Specifically, SSA fuses local and anchor attention outputs in the frequency domain to preserve visual quality and temporal consistency.
    \item Extensive experiments demonstrate that FreqForcing enables high-quality, minutes-long video generation, outperforming state-of-the-art training-free methods.
\end{itemize}

\section{Related Work}
\paragraph{AR video generation.}
Natively AR models~\cite{teng2025magi,chen2025skyreels,ruhe2024rolling,deng2025autoregressive} generate frames or video chunks sequentially under causal temporal dependencies, but typically require substantial computational resources for training. Distillation-based methods~\cite{yin2025slow,huang2026self,yang2025longlive,liu2025rolling,zhu2026causal} instead adapt a pretrained multi-step bidirectional teacher into a few-step causal student, inheriting strong visual priors, enabling real-time streaming generation and requiring fewer training resources. FreqForcing enhances the capability of AR video generation without additional training.
\vspace{-4.5mm}
\paragraph{Attention sink in language and video models.}
Attention sink~\cite{xiao2024efficient,gu2025attention} was first identified in AR language models, where a few initial tokens are retained in the rolling KV cache to preserve global context. Recent studies on AR video generation~\cite{li2026train,yang2025longlive,yi2025deep} similarly observe that attention sink plays a crucial role in stabilizing the video quality. In this paper, we illustrate the effectiveness of attention sink from various perspectives, showing that it partially preserves video quality during AR generation.
\vspace{-4.5mm}
\paragraph{Frequency-domain guidance for video generation.}
Frequency-domain guidance has been widely explored in video generation. FreeU~\cite{si2024freeu} rebalances low- and high-frequency features in the denoising U-Net without retraining. FreeInit~\cite{wu2024freeinit} iteratively refines the initial noise by retaining its low-frequency components while resampling the high-frequency ones, and FreqPrior~\cite{yuan2025freqprior} improves this frequency-filtered noise design to better preserve details and approximate a standard Gaussian distribution. FreeLong~\cite{lu2024freelong} utilizes SpectralBlend temporal attention to improve long-video generation. These methods, however, operate on bidirectional video generation models. FreqForcing instead performs frequency-domain fusion within AR self-rollout, enabling infinite-length video generation. 

\section{Preliminaries}
\paragraph{AR video diffusion models.}
AR video diffusion models predict the next frame conditioned on previously generated ones~\cite{yin2025slow,huang2026self,cui2025self,liu2025rolling,zhu2026causal}, with each prediction formulated as a diffusion denoising process. Let $\{t_0,t_1,\ldots,t_T\}$ denote the denoising schedule, where $t_0=0$ and $t_T=1000$. In the VAE-encoded latent space, the joint distribution of an $N$-chunk latent video is factorized as:
\begin{equation}
p_{\theta}\!\left(x_{t_0}^{[0,N)} \mid c\right)
=
p_{\theta}\!\left(x_{t_0}^{0} \mid c\right)
\prod_{i=1}^{N-1}
p_{\theta}\!\left(
x_{t_0}^{i}
\mid
x_{t_0}^{[0,i)},c
\right),
\end{equation}
where $c$ denotes the context. For the $i$-th latent chunk, the transition from noise level $t_j$ to $t_{j-1}$ is given by:
\begin{equation}
\begin{aligned}
x_{t_{j-1}}^{i}
&=
\Psi\!\left(
G_{\theta}\!\left(
x_{t_j}^{i},
t_j,
\mathbf{KV},
c
\right),
t_{j-1}
\right),
\end{aligned}
\label{eq:self-forcing}
\end{equation}
where $G_{\theta}$ is the denoising model, $\Psi$ denotes the forward diffusion operator, and $\mathbf{KV}$ denotes the sliding-window KV cache. The key-value pairs in $\mathbf{KV}$ are obtained through an additional model forward pass at timestep $t_0$~\cite{huang2026self,liu2025rolling,yang2025longlive,zhu2026causal}: \begin{equation}
\begin{aligned}
    (\mathbf{k}_i,\mathbf{v}_i)
&\leftarrow
G_{\theta}\!\left(
    x_0^{i},
    t_{0},
    \mathbf{KV},
    c
\right), \\
\mathbf{KV}
&=
\operatorname{Update}\!\left(
    \mathbf{KV},
    (\mathbf{k}_i,\mathbf{v}_i)
\right).
\end{aligned}
\label{eq:kv-cache-update}
\end{equation}

\paragraph{Self-Forcing.}
Teacher Forcing~\cite{jin2024pyramidal} and Diffusion Forcing~\cite{chen2024diffusion} train models on the ground-truth distribution, which leads to a train-test gap. By contrast, Self-Forcing~\cite{huang2026self} updates the KV cache with frames produced by the model itself during training. This design narrows the train-test gap and mitigates exposure bias. Self-Forcing further adopts a $4$-step denoising process which reduces sampling latency and enables real-time streaming video generation.

\begin{figure*}[t]
\centering
\includegraphics[width=\textwidth]{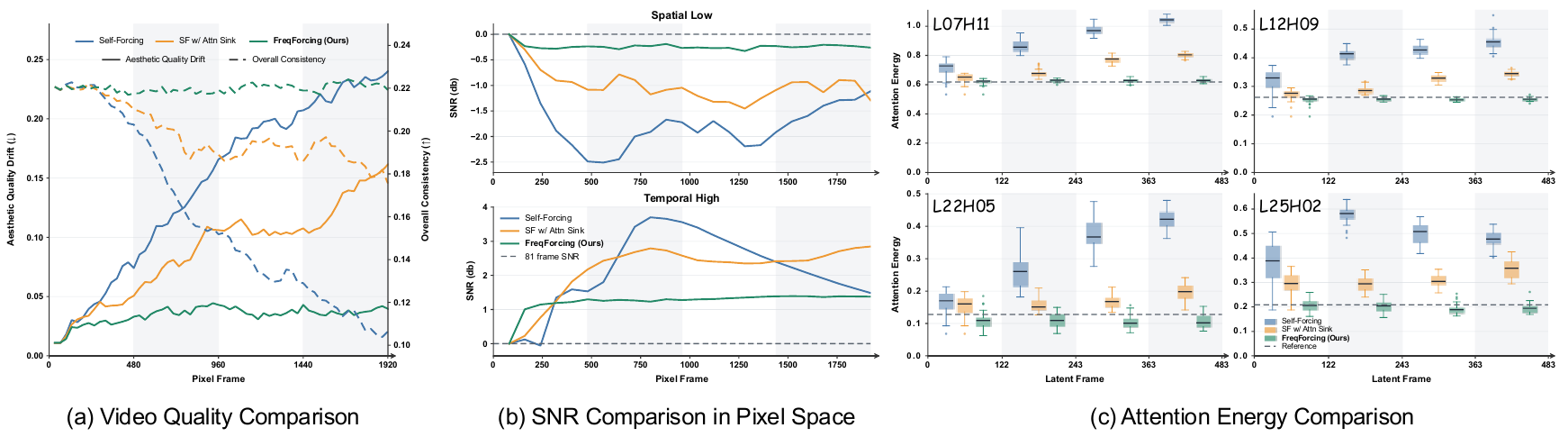}
\vspace{-6mm}
\caption{\textbf{Error analysis in AR generation.} \textbf{(a)}  Temporal evolution of aesthetic quality drift and overall consistency. \textbf{(b)} SNR of spatial low-frequency and temporal high-frequency components. \textbf{(c)} Distributions of attention output energy for selected heads from different DiT layers. FreqForcing maintains more stable visual quality, frequency-domain statistics, and attention output energy during inference.
}
\label{fig:error_analysis}
\end{figure*}

\section{Analysis of Error Accumulation}
\label{sec:error_analysis}
Recent works~\cite{yi2025deep,liu2025rolling} have observed that AR video generation suffers from error accumulation. Here, we provide a more comprehensive illustration and analyze the causes of visual degradation. 
\vspace{-4mm}
\paragraph{Visual quality evaluation.}
We first generate a set of long videos using vanilla Self-Forcing~\cite{huang2026self}.
To investigate the effectiveness of attention sink~\cite{xiao2024efficient}, we additionally evaluate Self-Forcing~\cite{huang2026self} equipped with attention sink.
Following Rolling Forcing~\cite{liu2025rolling}, we align the temporal indices of sink frames with the current attention window.
We quantify visual quality variations using aesthetic quality drift and overall consistency from VBench~\cite{huang2023vbench}.
As shown in Fig.~\ref{fig:error_analysis}(a), incorporating attention sink mitigates visual quality degradation.
Nevertheless, noticeable degradation persists as generation progresses.

Panels (a) and (b) of Fig.~\ref{fig:spectral_drift} further illustrate the visual degradation observed during long-horizon generation.
In particular, generated videos exhibit progressively pronounced color shifts and inconsistent subject appearances over time, indicating that neither inference strategy can effectively prevent visual degradation.
\vspace{-4mm}
\paragraph{Frequency-domain analysis.}
To systematically characterize how visual content evolves during generation, we analyze the frequency components of generated videos using the signal-to-noise ratio (SNR). We assume that frames generated within the pretrained horizon (\ie, the first 81 frames) maintain satisfactory visual quality and thus use them as high-quality references to evaluate subsequent frames' SNR. As shown in Fig.~\ref{fig:error_analysis}(b), vanilla Self-Forcing~\cite{huang2026self} exhibits a substantial decline in the SNR of spatial low-frequency components as generation progresses, corresponding to the pronounced color drift observed in pixel space. Meanwhile, the SNR of temporal high-frequency components initially increases and then decreases. This trend aligns with the observed transition from excessive temporal flickering to nearly static video content. Although incorporating attention sink improves overall visual quality, unnatural motion and camera transitions caused by error accumulation remain evident.
\vspace{-3mm}
\paragraph{Error propagation in attention computation.}
We perceive that the aforementioned SNR variations arise from imperfect attention outputs. Previously generated chunks are stored in the KV cache and continue to affect subsequent attention computation. For each query $j$ in chunk $i$, we denote the corresponding attention output by $A_{i,j}$ and then compute the mean-square energy $\mathcal{E}_{i}$ as: 
\begin{equation}
    \mathcal{E}_{i}
    =
    \frac{1}{d_h}
    \mathbb{E}_{j \sim \mathcal{U}(\mathcal{T}_i)}
    \left[
        \left\| A_{i,j} \right\|_2^2
    \right],
\end{equation}
where $\mathcal{T}_i$ denotes the set of tokens in the current chunk, with $|\mathcal{T}_i| = fhw$ and $d_h$ denotes the number of hidden dimensions for each attention head.

As illustrated in Fig.~\ref{fig:error_analysis}(c), during  vanilla Self-Forcing~\cite{huang2026self} inference, $\mathcal{E}_{i}$ expands significantly, ultimately leading to unsatisfactory predictions. Utilizing attention sink can mitigate this energy expansion to some extent. However, it still fails to prevent the visual degradation. 

These observations lead to two key insights: (1) Sink tokens help stabilize outputs by participating in attention computation. However, this implicit constraint remains insufficient to prevent error accumulation. (2) Errors propagate to subsequent generation through the imperfect key-value pairs, forming a positive feedback loop that progressively amplifies visual degradation. Without external correction, subsequent outputs are unable to recover the visual quality, ultimately leading to visual collapse.

\begin{figure*}[t]
\centering
\includegraphics[width=\textwidth]{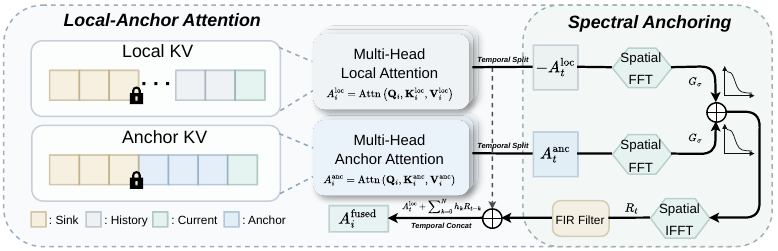}
\vspace{-6mm}
\caption{\textbf{Overview of FreqForcing.} FreqForcing rectifies attention outputs during autoregressive video generation via Spectral Self-Anchoring (SSA). SSA consists of three steps. \textbf{First}, before denoising the current chunk $i$, we cache high-quality key-value pairs for subsequent reference. \textbf{Second}, we design a dual-branch attention framework. \textbf{Third}, we fuse the high-frequency components of $A^{\mathrm{loc}}_i$ with the low-frequency components of $A^{\mathrm{anc}}_i$, yielding the final fused output $A^{\mathrm{fused}}_i$.
}
\label{fig:method}
\end{figure*}

\section{Method}
\label{sec:method}
Based on the above analysis, we introduce \textbf{FreqForcing}, a training-free framework that leverages Spectral Self-Anchoring (SSA) to correct attention outputs, enabling high-quality long-video generation.
Specifically, we modify the attention computation through three key steps.
First, before generating each chunk, we construct an anchor cache containing high-quality key-value pairs to provide stable references for subsequent inference.
Second, during inference, we employ a dual-branch attention mechanism to compute both local and anchor attention.
Third, we apply spectral anchoring to combine the high-frequency components of the local attention output with the low-frequency components of the anchor attention output, thereby improving visual quality while preserving temporal consistency.
Our framework is illustrated in Fig.~\ref{fig:method}.

\subsection{Local-Anchor Attention}
Our method maintains a local branch and an anchor branch during inference. The local branch uses conventional sliding-window spatiotemporal attention with sink tokens. The cached key-value pairs for the current chunk $i$ are organized as follows:
\begin{equation}
    \begin{aligned}
        \mathbf{K}^{\mathrm{loc}}_i
        &= [\mathbf{k}^{\mathrm{sink}} \parallel \mathbf{k}^{\mathrm{hist}}_i], \\
        \mathbf{V}^{\mathrm{loc}}_i
        &= [\mathbf{v}^{\mathrm{sink}} \parallel \mathbf{v}^{\mathrm{hist}}_i],
    \end{aligned}
\end{equation}
where $(\mathbf{k}^{\mathrm{sink}},\mathbf{v}^{\mathrm{sink}})$ are persistent sink tokens and $(\mathbf{k}^{\mathrm{hist}}_i,\mathbf{v}^{\mathrm{hist}}_i)$ contain the historical tokens maintained by a simple FIFO (First-In, First-Out) policy. The cache sizes for sink tokens and history tokens are denoted by $S$ and $H$, respectively. The local attention $A^{\mathrm{loc}}_i$ is calculated as follows:
\begin{equation}
    A^{\mathrm{loc}}_i = \operatorname{Softmax}\left(\frac{\mathbf{Q}_i [\mathbf{K}^{\mathrm{loc}}_i, \mathbf{k}_i]^{\top}}{\sqrt{d_h}}\right)[\mathbf{V}^{\mathrm{loc}}_i, \mathbf{v}_i],
\end{equation}
where $(\mathbf{k}_i,\mathbf{v}_i)$ are key-value pairs from the current chunk.

Once the clean latent $x_0^{i}$ is obtained, we update the local KV cache following the same update strategy as in Eq.~\ref{eq:kv-cache-update}, formulated as follows:
\begin{equation}
\begin{aligned}
    (\mathbf{k}_{i+1}^{\mathrm{hist}},\mathbf{v}_{i+1}^{\mathrm{hist}})
    &\leftarrow
    G_{\theta}\!\left(
        x_0^{i},
        t_{0},
        \mathbf{KV}^{\mathrm{loc}}_i,
        c
    \right), \\
    \mathbf{KV}^{\mathrm{loc}}_{i+1}
    &=
    \operatorname{Update}\!\left(
        \mathbf{KV}^{\mathrm{loc}}_i,
        (\mathbf{k}_{i+1}^{\mathrm{hist}},\mathbf{v}_{i+1}^{\mathrm{hist}})
    \right).
\end{aligned}
\label{eq:local-kv-cache-update}
\end{equation}
As discussed in Sec.~\ref{sec:error_analysis}, the KV cache update strategy in Eq.~\ref{eq:local-kv-cache-update} propagates accumulated errors into subsequent attention computations.
To mitigate this issue, we construct an anchor cache containing high-quality key-value pairs, which serves as a reliable reference for correcting errors introduced by local attention.

Specifically, the anchor cache consists of the following components:
\begin{equation}
    \begin{aligned}
        \mathbf{K}^{\mathrm{anc}}_i
        &= [\mathbf{k}^{\mathrm{sink}} \parallel
            \mathbf{k}^{\mathrm{anc}}_i], \\
        \mathbf{V}^{\mathrm{anc}}_i
        &= [\mathbf{v}^{\mathrm{sink}} \parallel
            \mathbf{v}^{\mathrm{anc}}_i],
    \end{aligned}
\end{equation}
where $(\mathbf{k}^{\mathrm{anc}}_i,\mathbf{v}^{\mathrm{anc}}_i)$ are obtained by recomputing the key-value pairs of the preceding chunk and the corresponding cache size is $M$. This recomputation uses only the sink tokens as cached context:
\begin{equation}
\begin{aligned}
    (\mathbf{k}^{\mathrm{anc}}_i,\mathbf{v}^{\mathrm{anc}}_i)
    &\leftarrow
    G_{\theta}\!\left(
        x_0^{i-1},
        t_0,
        \mathbf{kv}^{\mathrm{sink}},
        c
    \right), \\
    \mathbf{KV}^{\mathrm{anc}}_i
    &=
    \operatorname{Update}\!\left(
        \mathbf{KV}^{\mathrm{anc}}_{i-1},
        (\mathbf{k}^{\mathrm{anc}}_i,\mathbf{v}^{\mathrm{anc}}_i)
    \right).
\end{aligned}
\label{eq:anc-kv-cache-update}
\end{equation}
The anchor attention output is then computed as:
\begin{equation}
    A^{\mathrm{anc}}_i =
    \operatorname{Softmax}\left(
    \frac{\mathbf{Q}_i [\mathbf{K}^{\mathrm{anc}}_i, \mathbf{k}_i]^{\top}}{\sqrt{d_h}}
    \right)[\mathbf{V}^{\mathrm{anc}}_i, \mathbf{v}_i].
\end{equation}
Since $(\mathbf{k}^{\mathrm{anc}}_i,\mathbf{v}^{\mathrm{anc}}_i)$ are computed without relying on key-value pairs affected by accumulated errors, we regard $A^{\mathrm{anc}}_i$ as a high-quality reference for correcting the local attention.

\begin{table*}[t]
\centering
\small
\caption{\textbf{Quantitative comparison on VBench-Long~\cite{huang2023vbench} with Self-Forcing~\cite{huang2026self}.} All metrics are reported on \emph{60s} and \emph{120s} videos. Avg. Rank denotes the mean rank across the seven metrics. The best results are highlighted in \textbf{bold} with a darker blue background, and the second-best results are \underline{underlined} with a light blue background.}
\vspace{-2mm}
\begin{fullcelltabular}{\linewidth}{9}{l|ccccccc|c}
\toprule
Method & \shortstack{Dynamic\\Degree$\uparrow$}& \shortstack{Motion\\Smoothness$\uparrow$}& \shortstack{Overall\\Consistency$\uparrow$}& \shortstack{Imaging\\Quality$\uparrow$} & \shortstack{Aesthetic\\Quality$\uparrow$} & \shortstack{Subject\\Consistency$\uparrow$} & \shortstack{Background\\Consistency$\uparrow$} & \shortstack{Avg\\Rank$\downarrow$}\\
\midrule
\multicolumn{9}{c}{\emph{60s generation}} \\
\midrule
Self-Forcing~\cite{huang2026self} & 32.96 & \secondbest{98.18} & 18.95 & 66.05 & 56.34 & 96.85 & \secondbest{96.29} & 3.93 \\
\midrule
Infinity-RoPE~\cite{yesiltepe2026infinity} & 56.20 & 97.81 & 18.23 & \secondbest{69.46} & 58.08 & 96.52 & 95.54 & 4.14 \\
Deep Forcing~\cite{yi2025deep} & \secondbest{62.58} & 98.08 & 20.28 & 68.41 & \secondbest{59.89} & 96.92 & 96.21 & 3.00 \\
MemRoPE~\cite{kim2026memrope} & 54.64 & \best{98.23} & \secondbest{20.41} & \best{69.62} & 58.98 & \secondbest{97.19} & \secondbest{96.29} & \secondbest{2.21} \\
\textbf{FreqForcing (Ours)} & \best{64.89} & 98.05 & \best{21.44} & 69.39 & \best{61.25} & \best{97.49} & \best{96.47} & \best{1.71} \\
\midrule
\multicolumn{9}{c}{\emph{120s generation}} \\
\midrule
Self-Forcing~\cite{huang2026self} & 27.37 & \secondbest{98.08} & 16.28 & 61.47 & 51.16 & 96.32 & \secondbest{96.24} & 4.14 \\
\midrule
Infinity-RoPE~\cite{yesiltepe2026infinity} & 54.71 & 97.72 & 17.90 & \secondbest{69.17} & 56.98 & 96.51 & 95.50 & 3.86 \\
Deep Forcing~\cite{yi2025deep} & \secondbest{64.41} & 97.92 & \secondbest{20.15} & 68.85 & \secondbest{59.43} & 96.74 & 96.10 & 3.00 \\
MemRoPE~\cite{kim2026memrope} & 54.07 & \best{98.20} & 20.02 & 69.12 & 57.57 & \secondbest{96.99} & 96.15 & \secondbest{2.71} \\
\textbf{FreqForcing (Ours)} & \best{65.41} & 98.00 & \best{21.45} & \best{69.39} & \best{61.02} & \best{97.46} & \best{96.41} & \best{1.29} \\
\bottomrule
\end{fullcelltabular}
\label{tab:SF_quantitative}
\end{table*}

\begin{table*}[htbp]
\centering
\small
\caption{\textbf{Quantitative comparison on VBench-Long~\cite{huang2023vbench} with Causal Forcing~\cite{zhu2026causal}.} All metrics are reported on \emph{120s} generations. The best results are highlighted in \textbf{bold} with a darker blue background, and the second-best results are \underline{underlined} with a light blue background.}
\vspace{-2mm}
\begin{fullcelltabular}{\linewidth}{9}{l|ccccccc|c}
\toprule
Method & \shortstack{Dynamic\\Degree$\uparrow$}& \shortstack{Motion\\Smoothness$\uparrow$}& \shortstack{Overall\\Consistency$\uparrow$}& \shortstack{Imaging\\Quality$\uparrow$} & \shortstack{Aesthetic\\Quality$\uparrow$} & \shortstack{Subject\\Consistency$\uparrow$} & \shortstack{Background\\Consistency$\uparrow$} & \shortstack{Avg\\Rank$\downarrow$}\\
\midrule
\multicolumn{9}{c}{\emph{120s generation}} \\
\midrule
Causal Forcing~\cite{zhu2026causal} & 62.09 & 95.91 & 14.77 & 59.67 & 46.99 & 94.30 & \secondbest{95.47} & 4.57 \\
\midrule
Infinity-RoPE~\cite{yesiltepe2026infinity} & \secondbest{83.18} & 95.99 & 19.44 & \secondbest{69.38} & \secondbest{57.63} & \secondbest{95.42} & 95.19 & \secondbest{2.57} \\
Deep Forcing~\cite{yi2025deep} & 81.92 & \secondbest{96.78} & 18.34 & 68.52 & 55.97 & 95.23 & 95.17 & 3.29 \\
MemRoPE~\cite{kim2026memrope} & \best{87.92} & 96.27 & \secondbest{20.02} & 69.10 & 55.82 & 95.06 & 95.03 & 3.14 \\
\textbf{FreqForcing (Ours)} & 80.32 & \best{97.17} & \best{20.72} & \best{69.92} & \best{59.84} & \best{96.87} & \best{96.11} & \best{1.43} \\
\bottomrule
\end{fullcelltabular}
\vspace{-3mm}
\label{tab:CF_quantitative}
\end{table*}

\subsection{Spectral Anchoring}
\label{sec:spectral-anchoring}
We apply spectral anchoring to correct local attention outputs on a per-frame basis. Let $f$ denote the number of latent frames per chunk. The $r$-th latent frame in chunk $i$ has the global index $t=if+r$, where $r\in\{0,\ldots,f-1\}$. We denote its local and anchor attention outputs by $A_t^{\mathrm{loc}}$ and $A_t^{\mathrm{anc}}$, respectively, and define the correction residual as follows:
\begin{equation}
    D_t =
        A_t^{\mathrm{anc}} - A_t^{\mathrm{loc}}.
\end{equation}
To preserve the high-frequency components of the local attention output, we apply corrections only to its low-frequency components:
\begin{equation}
    R_t
    =
    \mathcal{F}^{-1}\!\left[
        G_\sigma \odot \mathcal{F}(D_t)
    \right],
\end{equation}
where $\mathcal{F}$ and $\mathcal{F}^{-1}$ denote the spatial Fast Fourier Transform and its inverse operation,
respectively, and $G_\sigma$ is the spatial Gaussian low-pass filter.

To ensure temporal consistency, we apply an $N$-th order causal finite impulse response (FIR) filter along the temporal dimension:
\begin{equation}
    \widehat{R}_t
    =
    \sum_{k=0}^{N} h_k R_{t-k},
\end{equation}
where $h_k$ denotes the $k$-th filter coefficient. The FIR filter incorporates past correction terms, thereby smoothing the final output.
The final fused attention output is:
\begin{equation}
    A_t^{\mathrm{fused}}
    =
    A_t^{\mathrm{loc}} + \widehat{R}_t.
\end{equation}
As shown in Fig.~\ref{fig:error_analysis}, spectral anchoring maintains stable attention output energy over long horizons. The generated videos exhibit minimal color drift or motion stagnation, while corresponding VBench~\cite{huang2023vbench} scores remain nearly constant throughout extended generation.

\subsection{Temporal RoPE Alignment}
Modern video diffusion transformers (DiTs)~\cite{wan2025wan} commonly employ 3D rotary positional embeddings (3D-RoPE) to encode spatiotemporal positions. During AR generation, the temporal indices of newly generated frames continuously increase. This can cause the temporal offsets between the current query tokens and the sink tokens to exceed the training range. To address this issue, inspired by prior works~\cite{liu2025rolling,kim2026memrope}, we store the key tokens without applying RoPE. Before attention, we dynamically assign their temporal indices to right before the current chunk. This strategy mitigates artifacts caused by positional extrapolation.

\begin{figure*}[t]
\centering
\includegraphics[width=\linewidth]{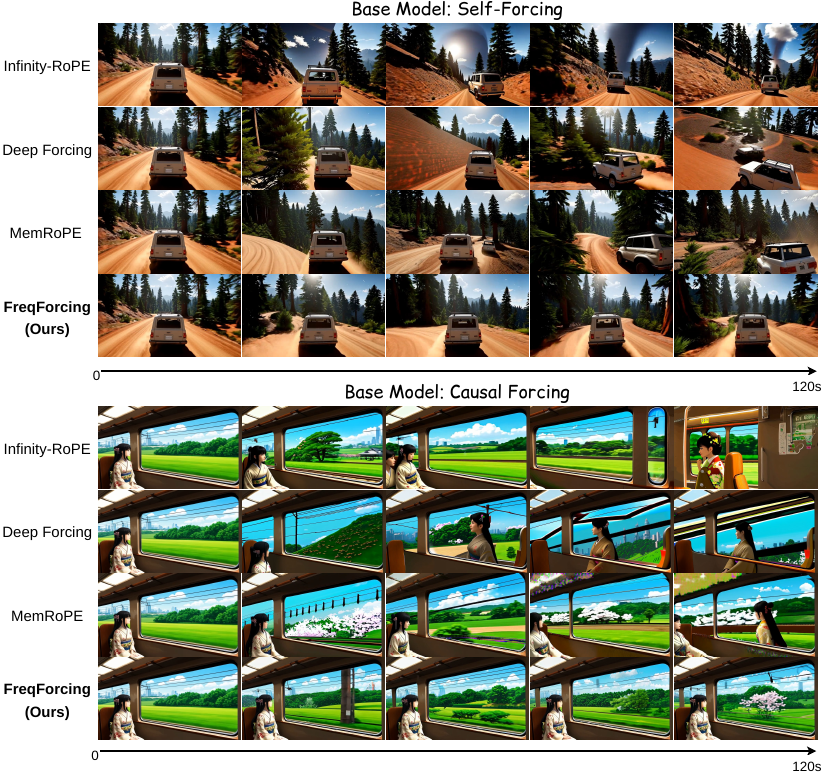}
\vspace{-9mm}
\caption{\textbf{Qualitative comparisons on \textit{120s} videos.} We compare FreqForcing with representative AR video diffusion methods.}

\label{fig:methods comparison}
\end{figure*}

\section{Experiments}
\subsection{Experimental Settings}
\paragraph{Implementation details.} We evaluate FreqForcing on chunk-wise Self-Forcing~\cite{huang2026self} and Causal Forcing~\cite{zhu2026causal} built on Wan2.1-T2V-1.3B~\cite{wan2025wan}, with chunk size $f=3$. Both models generate videos autoregressively with a 4-step denoising schedule at timesteps $\left \{1000, 750, 500, 250\right \}$. Anchor attention computation and spectral anchoring are performed only during the first two denoising steps. The cache sizes for sink, history, and anchor tokens are set to $S=3$, $H=6$, and $M=3$, respectively. For spectral anchoring, we use a Gaussian low-pass filter with a standard deviation $\sigma = 0.125$, and a second-order causal FIR filter to incorporate past correction terms.
\vspace{-5mm}
\paragraph{Evaluation.} To evaluate the performance of long-video generation, we use the first \textit{128} prompts from MovieGen~\cite{polyak2024movie} under the same prompt selection protocol as Self-Forcing++~\cite{cui2025self}, with each prompt refined by Qwen/Qwen2.5-7B-Instruct~\cite{yang2025qwen3} as in Self-Forcing~\cite{huang2026self}. We report seven VBench-Long~\cite{huang2023vbench} metrics to comprehensively evaluate our method for \textit{60s} and \textit{120s} videos, with a resolution of $832\times 480$ at 16 fps. We further evaluate five aspects of video quality through a user preference study following the 2AFC protocol~\cite{kim2026memrope,yi2025deep}.

\subsection{Results}
\paragraph{Comparison.}
We compare our proposed FreqForcing against several state-of-the-art AR long video generation methods, including Infinity-RoPE~\cite{yesiltepe2026infinity}, Deep Forcing~\cite{yi2025deep}, and MemRoPE~\cite{kim2026memrope}. All methods can be applied to both Self-Forcing~\cite{huang2026self} and Causal Forcing~\cite{zhu2026causal} in a training-free manner. We generate 60s and 120s videos to evaluate the performance of different methods quantitatively and qualitatively. The results are in Tab.~\ref{tab:SF_quantitative}, Tab.~\ref{tab:CF_quantitative}, and Fig.~\ref{fig:methods comparison}. 
\begin{table*}[t]
\centering
\small
\caption{\textbf{Ablation on proposed components.} All configurations generate \emph{60s} videos with the same local KV cache size (\ie, 12 latent frames). Specifically, we isolate the contributions of the proposed anchor attention and spectral anchoring.}
\vspace{-1.8mm}
\begin{fullcelltabular}{\linewidth}{8}{l|ccccccc}
\toprule
Configuration & \shortstack{Dynamic\\Degree$\uparrow$} & \shortstack{Motion\\Smoothness$\uparrow$} & \shortstack{Overall\\Consistency$\uparrow$} & \shortstack{Imaging\\Quality$\uparrow$} & \shortstack{Aesthetic\\Quality$\uparrow$} & \shortstack{Subject\\Consistency$\uparrow$} & \shortstack{Background\\Consistency$\uparrow$} \\
\midrule
Self-Forcing~\cite{huang2026self} (Baseline) & 32.96 & \best{98.18} & 18.95 & 66.05 & 56.34 & 96.85 & 96.29 \\
Self-Forcing~\cite{huang2026self} w/ Attention Sink & 56.88 & \best{98.18} & 20.29 & 68.43 & 58.17 & 97.00 & 96.16 \\
\textbf{FreqForcing (w/o FIR filter)} & 64.67 & 97.82 & \secondbest{21.42} & \secondbest{69.34} & \secondbest{61.09} & \best{97.51} & \best{96.50} \\
\textbf{FreqForcing (w/ high-pass filter)} & \best{67.87} & 97.74 & 20.84 & 68.17 & 59.94 & 96.85 & 96.16 \\
\midrule
\textbf{FreqForcing (w/ low-pass filter)} & \secondbest{64.89} & \secondbest{98.05} & \best{21.44} & \best{69.39} & \best{61.25} & \secondbest{97.49} & \secondbest{96.47} \\
\bottomrule
\end{fullcelltabular}
\vspace{-1mm}
\label{tab:ablation}
\end{table*}
\vspace{-4mm}
\paragraph{Quantitative results.} Table~\ref{tab:SF_quantitative} reports the quantitative results on Self-Forcing~\cite{huang2026self} for 60s and 120s video generation. FreqForcing achieves the best average rank at both durations.
Notably, vanilla Self-Forcing~\cite{huang2026self} performs well in motion smoothness and background consistency, primarily because it tends to generate nearly static and unusable content during long-video generation.
This observation highlights that achieving a high score on an individual VBench~\cite{huang2023vbench} metric does not necessarily indicate superior video quality, emphasizing the importance of evaluations across multiple metrics.
Furthermore, FreqForcing exhibits only marginal performance degradation when extending the generation duration from 60s to 120s, confirming its ability to maintain high video quality. Results for 120s generation on Causal Forcing~\cite{zhu2026causal} are shown in Tab.~\ref{tab:CF_quantitative}. FreqForcing achieves the best average rank (1.43), which demonstrates its transferability across different autoregressive video diffusion models.
\vspace{-4mm}
\paragraph{Qualitative results.} 
Figure~\ref{fig:methods comparison} presents qualitative comparisons of 120s videos generated by different methods based on Self-Forcing~\cite{huang2026self} and Causal Forcing~\cite{zhu2026causal}.
While Infinity-RoPE~\cite{yesiltepe2026infinity}, Deep Forcing~\cite{yi2025deep}, and MemRoPE~\cite{kim2026memrope} partially mitigate visual degradation, their generated videos still suffer from noticeable color shifts and inconsistencies in subject appearance.
In contrast, FreqForcing maintains stable visual quality, consistent subject appearance, and natural motion dynamics throughout long-horizon generation.
\vspace{-7.5mm}
\paragraph{User study.} To assess human preferences, we conducted a user study with 24 participants following the 2AFC protocol. Each participant was required to compare 24 pairs of 120s videos, evaluating FreqForcing against other training-free methods for long video generation on both Self-Forcing~\cite{huang2026self} and Causal Forcing~\cite{zhu2026causal}. Participants assessed five criteria: color consistency, subject consistency, background consistency, text alignment, and overall preference. As shown in Tab.~\ref{tab:user_study}, participants preferred FreqForcing across all five criteria on both base models, consistent with the previous quantitative results in Tab~\ref{tab:SF_quantitative} and ~\ref{tab:CF_quantitative}.

\begin{table}[t]
\centering
\footnotesize
\caption{\textbf{User study results (\%).} Values indicate the percentage of votes favoring FreqForcing over the baseline methods.}
\label{tab:user_study}
\vspace{-2mm}
\begin{fullcelltabular}{\linewidth}{6}{lccccc}
\toprule
Baseline & \shortstack{Color\\Cons.} & \shortstack{Subject\\Cons.} & \shortstack{Bg.\\Cons.} & \shortstack{Text\\Align.} & \shortstack{Overall\\Pref.} \\
\midrule
\multicolumn{6}{l}{\textbf{\textit{Results on Self-Forcing}}~\cite{huang2026self}} \\
\midrule
Self-Forcing~\cite{huang2026self} &  97.22 &  98.61&  91.67&  98.61&  100.00 \\
Infinity-RoPE~\cite{yesiltepe2026infinity} & 93.06 & 91.67 & 97.22 & 80.56 & 72.22  \\
Deep Forcing~\cite{yi2025deep} & 88.89 & 91.67 & 93.06 & 83.33 & 91.67 \\
MemRoPE~\cite{kim2026memrope} & 91.67 & 88.89 & 84.72 & 86.11 & 86.11 \\
\midrule
\multicolumn{6}{l}{\textbf{\textit{Results on Causal Forcing}}~\cite{zhu2026causal}} \\
\midrule
Causal Forcing~\cite{zhu2026causal} & 97.22 & 100.00 & 95.83 & 91.67 & 100.00  \\
Infinity-RoPE~\cite{yesiltepe2026infinity} & 84.72 & 81.94 & 76.39 & 95.83 & 86.11  \\
Deep Forcing~\cite{yi2025deep} & 93.06 & 98.61 & 98.61 & 87.50 & 84.72 \\
MemRoPE~\cite{kim2026memrope} & 75.00 & 76.39 & 79.17 & 84.72 & 73.61 \\
\bottomrule
\end{fullcelltabular}
\end{table}
\vspace{-3mm}
\paragraph{Ablation study.} As illustrated in Tab.~~\ref{tab:ablation}, we isolate the contributions of the proposed components. \textbf{First}, we compare FreqForcing with Self-Forcing~\cite{huang2026self} equipped with attention sink to assess the benefits of anchor-guided correction. The results show that incorporating anchor attention substantially improves the dynamic degree, aesthetic quality, and visual consistency. \textbf{Second}, we demonstrate the impact of the causal FIR filter proposed in Sec.~\ref{sec:spectral-anchoring}. Quantitative results indicate that incorporating the FIR filter further improves temporal consistency and overall quality. \textbf{Third}, we replace the spatial low-pass filter with a high-pass filter to investigate the importance of applying correction to low-frequency components. With the low-pass filter, FreqForcing achieves better scores on motion smoothness and consistency metrics, supporting our choice to apply corrections to low-frequency components.

\vspace{-4mm}
\paragraph{Efficiency.}\begin{wraptable}[5]{r}{0.4\columnwidth}
\vspace{-4mm}
\centering
\footnotesize
\setlength{\tabcolsep}{4pt}
\captionsetup{skip=3pt}
\caption{\textbf{Inference speed.}}
\label{tab:inference_efficiency}
\begin{tabular}{@{}lc@{}}
\toprule
Method & FPS \\
\midrule
Self-Forcing~\cite{huang2026self} & 2.978 \\
+ FreqForcing & 3.040 \\
\bottomrule
\end{tabular}
\vspace{-2mm}
\end{wraptable}
FreqForcing introduces computational overhead from three sources: \textbf{(a)} an auxiliary model forward pass for updating the anchor cache, \textbf{(b)} anchor attention computation, and \textbf{(c)} spectral anchoring for corrections.

We mitigate this overhead through two strategies.
\textbf{First}, despite employing dual-branch attention, the overall window size of the local and anchor branches remains identical to that of Self-Forcing~\cite{huang2026self} (\ie, 21 latent frames). \textbf{Second}, Spectral Self-Anchoring is applied only during the first two denoising steps. For the remaining denoising steps and KV cache updating step, only local attention is calculated. Following MemRoPE~\cite{kim2026memrope}, we measure inference speed on a single NVIDIA A6000 GPU. As shown in Tab.~\ref{tab:inference_efficiency}, FreqForcing achieves comparable inference speed to vanilla Self-Forcing~\cite{huang2026self}, demonstrating its efficiency.

\section{Conclusion}
We introduce FreqForcing, a training-free framework for AR long-video generation. We propose Spectral Self-Anchoring to compensate for error accumulation by correcting attention outputs. Extensive quantitative evaluations and frame-by-frame qualitative comparisons demonstrate that FreqForcing achieves state-of-the-art performance in long-video generation. Furthermore, we demonstrate its transferability across different AR video diffusion models. FreqForcing provides a simple yet effective solution for high-quality long-video generation.

{
    \small
    \bibliographystyle{ieeenat_fullname}
    \bibliography{main}
}


\end{document}